\documentclass[twocolumn,12pt]{IEEEtran}
\usepackage{booktabs}
\usepackage{adjustbox}
\usepackage{amsmath,graphicx}
\usepackage{amsmath,amssymb}
\usepackage{amsthm}
\usepackage{myMacros}
\usepackage{amsmath,amssymb}
\usepackage{algpseudocode}
\usepackage{algorithm}
\usepackage{float}
\usepackage{color}
\usepackage{graphicx}
\usepackage{epstopdf}
\usepackage{epsfig}
\usepackage{icomma}
\usepackage{subcaption}
\usepackage{colortbl}
\usepackage{hhline}
\usepackage{algorithmicx}
\usepackage{balance}
\usepackage{multicol}
\usepackage{soul}
\usepackage{xcolor}
\usepackage{etoolbox} 

\BeforeBeginEnvironment{equation}{\vspace{-3pt}}
\AfterEndEnvironment{equation}{\vspace{-3pt}}

\usepackage{mathtools}
\usepackage[top=0.90in, bottom=1.0in, left=0.60in, right=0.60in]{geometry}

\usepackage{cite}

\ifCLASSINFOpdf
\else
\fi
\begin{document}
%

\title{Flexible Payload Configuration for Satellites using Machine Learning}
%
%
%

\author{\IEEEauthorblockN{ Marcele O. K. Mendon\c{c}a, Flor G. Ortiz-Gomez, Jorge Querol,
       Eva Lagunas, \\ Juan A. Vásquez Peralvo, Victor Monzon Baeza, Symeon Chatzinotas and
        Bjorn Ottersten}   \\
\IEEEauthorblockA{\normalsize{\textit{Interdisciplinary Centre for Security Reliability and Trust (SnT)
 - University of Luxembourg, Luxembourg. 
Corresponding Author:  marcele.kuhfuss@uni.lu }}}

}

\maketitle
\thispagestyle{empty}
\begin{abstract}

Satellite communications, essential for modern connectivity, extend access to maritime, aeronautical, and remote areas where terrestrial networks are unfeasible. Current GEO systems distribute power and bandwidth uniformly across beams using multi-beam footprints with fractional frequency reuse. However, recent research reveals the limitations of this approach in heterogeneous traffic scenarios, leading to inefficiencies. To address this, 
this paper presents a machine learning (ML)-based approach to Radio Resource Management (RRM).

We treat the RRM task  as a regression ML problem, integrating RRM objectives  and constraints into the loss function that the ML algorithm aims at minimizing.   Moreover, we introduce a context-aware ML metric that evaluates the ML model's performance but also considers the impact of its resource allocation decisions on the overall performance of the communication system.

\end{abstract}


%
\IEEEpeerreviewmaketitle

\setlength\doublerulesep{0.7pt}

\section{Introduction}
\label{sec:intro}

Satellite networks offer an appealing solution for delivering ubiquitous connectivity across diverse domains such as the maritime and aeronautical markets and communication services to remote regions \cite{sharma2018satellite}. Current Geostationary (GEO) broadband satellite systems use a multibeam footprint strategy to enhance spectrum utilization.  In these systems, both power and bandwidth resources are typically allocated uniformly across the various beams. While this uniform allocation simplifies resource management, it may lead to inefficiencies in scenarios with varying traffic demands. Some beams may experience high demand, exceeding their available capacity, while others may have underutilized resources. This challenge has prompted research into more adaptive and dynamic resource allocation methods. In this regard, flexible payloads have emerged as an enabling technology to manage limited satellite resources by dynamically adapting the frequency, bandwidth, and power of the payload transponders according to users' demand \cite{kisseleff2020radio}.

Existing approaches aim to minimize the difference between offered and required capacity while adding constraints in terms of power \cite{choi2005optimum,liu2020ag}, and co-channel interference \cite{cocco2017radio}. The power allocation derived in \cite{choi2005optimum} is solved using water-filling, whereas a sub-optimal complexity game-based dynamic power allocation (AG-DPA) solution is proposed in \cite{liu2020ag}.  
A modified simulated annealing algorithm, as presented in \cite{cocco2017radio}, outperforms conventional payload designs in matching requested capacity across beams, emphasizing its effectiveness. However, the intricate computational complexities associated with these algorithms can significantly limit their practical applicability within real-world systems. Moreover, these approaches do not adequately consider the dynamic nature of capacity requests that change over time.  In this context, Machine learning (ML) algorithms emerge as a more favorable alternative, as they are able to learn from varying capacity request scenarios. 

ML algorithms have gained popularity in satellite communications, particularly in resource allocation \cite{ferreira2018multiobjective}.  Some studies explored reinforcement learning (RL) techniques \cite{huang2023sequential} to cope with the time-varying capacity; however, they introduced additional delays due to online payload controller training. Also, the RL exploration phase, aimed at discovering optimal strategies through action exploration, can occasionally result in system outages or disruptions when untested actions are selected. In contrast, \cite{ortiz2022machine} adopts a multi-objective optimization approach using supervised learning, offering an alternative perspective. 

In this work, we extend the ML-based method in \cite{ortiz2022machine} which originally employed a convolutional neural network (CNN) for solving the RRM task as a classification problem. In this approach, the ML model's objective is to select the best payload configuration from a discrete set of power and bandwidth combinations, treated as distinct classes. This technique considers 8 beams with 12 configurations each, giving a total of $4.3 \times 10^8$ potential payload configurations. We expand to 10 beams with 9 configurations each, totaling $3.5 \times 10^9$ configurations. Although the number of configurations decreases after applying the system constraints, incorporating more beams inevitably increases the number of classes. Having many classes complicates the ML model evaluation as traditional metrics like accuracy can be inadequate, and metrics like recall may not fully depict system performance. The situation worsens when dealing with imbalanced class distributions since the models may favor dominant classes, leading to bias. To address this, we reframe the RRM task as a regression problem, incorporating RRM objectives and constraints into the ML loss function.   We also introduce a new metric to assess the ML model's performance, offering an alternative and insightful way to evaluate its effectiveness in the context of RRM.

This paper is organized as follows.  Section \ref{sec:system} introduces the flexible payload architecture and outlines the RRM task. Section \ref{sec:proposed} compares regression and classification-based ML methods for flexible payload. In Section \ref{sec:metrics}, we present metrics for evaluating model performance, including a new ML metric for RRM. The methods are evaluated  in Section \ref{sec:sim_res}. Finally some conclusion remarks are included in Section \ref{sec:concc}.


\section{System Model and Problem Formulation} \label{sec:system}

We consider a GEO satellite system with a single multi-beam GEO satellite that covers a wide Earth region via $B$ spot-beams.  We focus on the forward link, considering $U$ single-antenna user terminals (UTs) distributed across the satellite's coverage area.  We assume that the considered payload can adaptably handle per-beam power and bandwidth resources.

\subsection{Link-budget analysis}\label{subsec:link}

The offered capacity  $C_b$ can be written as
\begin{equation}\label{eq:capacity_off}
    C_b  = \text{BW}_b \cdot \text{SE}_b,
\end{equation}
where $\text{SE}_b$ is the spectral efficiency (SE) for beam $b$ in bps/Hz \cite{ortiz2022supervised}. The SE is a function of the  carrier to interference plus noise ratio (CINR) in the $b$-th beam $\text{CINR}_b$. 
The CINR in dB can be written as
\begin{equation}
    10^{\frac{-\text{CINR}_b}{10}} = 10^{\frac{-\text{CIR}_b}{10}} + 10^{\frac{-\text{CNR}_b}{10}},
\end{equation}
where $\text{CIR}_b$ is the carrier to interference ratio and $\text{CNR}_b$ is the carrier to noise ratio for beam $b$ in dB. The CIR represents the ratio of the power allocated at $b$-th beam ($P_b$, in dBW) to the interference power at $b$-th beam ($I_b$, in dBW). The CNR (in dB) can be obtained 
\begin{equation}
    \text{CNR}_b = \text{EIRP3dB}_b  + G/T - A - k -   \text{BW}_b,
\end{equation}
where $\text{EIRP3dB}_b = P_b + G_b$ is the effective isotropic radiated power in dBW,  $G_b$ is the beam gain that depends on the half power beamwidth $\theta_{\rm 3dB}$ in dBi, $G/T$ is the merit figure of the user terminal, $A$ is the free space attenuation in clear sky conditions in dB, and $k$ is the  Boltzmann constant. 

With a particular $P_b$, we determine the $\text{CINR}_b$ and subsequently the $\text{SE}_b$. Then, we obtain the capacity as in equation (\ref{eq:capacity_off}) using $\text{SE}_b$ and $\text{BW}_b$.


\subsection{Traffic demand}\label{subsec:trafic_model}

To generate instances of satellite traffic demand at specific time instances, we employ the SnT Traffic Emulator \cite{al2020traffic}. This emulator utilizes three distinct input datasets: population data, aeronautical data, and maritime data. The data are processed to create a matrix representing the traffic demand. In this matrix, each position $i, j$ corresponds to a geographic location, and the value $r_{i,j}$ denotes the traffic demand in that specific geographic location in bits per second (bps). From $r_{i,j}$, the requested capacity $R_b$ is calculated by aggregating all the $(i,j)$ points within the coverage region of beam  $b$. 


\subsection{RRM task }\label{subsec:rrm_loss}
The RRM aims to effectively allocate the available satellite resources such as
power $P_b$ and bandwidth $\text{BW}_{b_c}$ so that  $C_b$ matches $R_b$ for each beam $b = 1, \cdots B$ over time $t$, avoiding resource waste. The RRM task can be formulated as the following minimization problem \cite{ortiz2022supervised}
\begin{equation} \label{eq:opt_fun}
\begin{aligned}
 \min_{P_b(t), \text{BW}_{b_c}(t)} \quad  & \frac{\beta_1}{B} \sum_{b=1}^B |C_b(t) - R_b(t)| + \\
+ \frac{\beta_2}{B} \sum_{b=1}^B P_b(t)  & + \frac{\beta_3}{B}\sum_{c=1}^{N_c} \sum_{b_c=1}^{B_c}  \text{BW}_{b_c}(t)  
\end{aligned}
\end{equation}

\begin{alignat}{2}
\text{s.t: } ~~&   C_b(t) \geq R_b(t) ~ \text{if}  ~ \{ P_b(t) < P_{\text{max},b} \nonumber \\
~~& \text{and } ~\text{BW}_{b_c} < \text{BW}_{\text{max}, b} \}
  \label{eq:opt_fun_constraint1} \\
                     ~~&  C_b(t)  = C_{\rm max}(t) ~ \text{if}  ~\{ P_b(t) = P_{\text{max},b} \nonumber  \\
                     ~~& \text{and } ~\text{BW}_{b_c} = \text{BW}_{\text{max}, b} \} \label{eq:opt_fun_constraint2} \\
                    ~~&  \sum_{b=1}^B P_b(t) \leq P_{\rm \text{max},T} \label{eq:opt_fun_constraint3} \\
                     ~~& \sum_{b_c=1}^{B_c} \text{BW}_{b_c} \leq \text{BW}_{\text{max}, c}.\label{eq:opt_fun_constraint4} 
\end{alignat}

The cost function in equation (\ref{eq:opt_fun}) aims to  simultaneously minimize three terms:
the difference between $C_b$ and $R_b$, the power $P_b$ (in W), and the bandwidth $\text{BW}_{b_c}$ (in Hz) across all beams for each time instant $t$. The weights $\beta_1$, $\beta_2$, and $\beta_3$ are assigned to each term to indicate its relative importance.

The constraint in equation (\ref{eq:opt_fun_constraint1}) ensures that offered capacity either meet or surpass the required capacity for each beam, under the condition that both the power and bandwidth allocations for the $b$-th beam at time $t$ do not exceed their upper bounds. On the other hand, if the required capacity  in the $b$-th beam is greater than the system can provide, the offered capacity within the $b$-th beam is inevitably capped at its maximum attainable value as accounted in (\ref{eq:opt_fun_constraint2}). 

To manage the overall power consumption, the total power $\sum_{b=1}^B P_b(t)$ is constrained  to not surpass the prescribed upper limit $P_{\rm \text{max},T}$ in equation (\ref{eq:opt_fun_constraint3}). Moreover, the constraint in equation (\ref{eq:opt_fun_constraint4})  imposes an upper limit on the  total bandwidth allocated in each color of the frequency plan, ensuring that it does not exceed the available bandwidth per color, $\text{BW}_{\text{max}, c}$.  The total bandwidth is allocated to the beams of each color $c$ within the frequency plan comprising $N_c$ colors where $B_c$ is the number of beams with the same frequency and polarization defined by color $c$.

\section{Proposed ML-based flexible payload methods}\label{sec:proposed}


In this section, we propose a CNN model to solve the RRM task as a regression problem, which involves determining the optimal payload configuration for specific traffic demands. Subsection \ref{subsec:dataset} introduces the dataset used to train and evaluate the model. In subsection \ref{subsec:classification}, we summarize the CNN model used for classification in \cite{ortiz2022machine} and in subsection \ref{subsec:regression} we detailed our proposed model.


\subsection{Dataset and preprocessing}\label{subsec:dataset}

The dataset consists of $M$ labeled examples, where each example is a data point with associated features and a target label. 

The data points are matrices representing the traffic demand at each geographic location in the service area. Preprocessing steps include data reduction via Max-Pooling filters, standardization, and principal component analysis (PCA) to extract relevant features and reduce training complexity.

The target label corresponds to the optimal payload configuration for a given traffic demand matrix. This configuration minimizes a cost function (equation \ref{eq:opt_fun}) while satisfying the constraints (equations \ref{eq:opt_fun_constraint1}-\ref{eq:opt_fun_constraint4}). The nature of the target label varies depending on whether it's a classification or regression problem. In classification, the label is categorical, denoting the class to which each data point belongs. In regression, it's continuous, representing numeric values. We outline the specific target labels for each ML problem as follows.

\subsection{ML-based flexible payload via classification}\label{subsec:classification}
In \cite{ortiz2022machine}, the RRM task is treated as a classification problem with $L$ classes representing possible payload configurations. The softmax activation function is applied in the CNN's output layer to convert raw output scores (logits) into a probability distribution over these classes. The final class is obtained by selecting the class with the highest probability. The model is trained by minimizing cross-entropy error, a standard loss function for classification tasks that measures the dissimilarity between predicted class probabilities and actual labels.

In this classification approach, the original optimization problem (as represented in equation (\ref{eq:opt_fun})) is only considered when generating the training dataset and defining the classes. That is, the optimization problem is not considered when formulating the loss function used to train the neural network. In essence, the loss function used during the neural network's training phase is designed to optimize the model's performance and is distinct from the original optimization problem associated with RRM. 

\subsection{ML-based flexible payload via regression}\label{subsec:regression}
When approaching the RRM task as a regression problem, the target labels are the desired offered capacity values for each beam. The linear activation function is used so that the neurons produce continuous values as their outputs: the predicted offered capacity values. In this case, the payload configuration with offered capacity values more similar to the ones obtained by the ML-model is selected. The similarity here is measured in terms of minimum mean absolute error (MAE).

 The model is trained by minimizing the mean square error (MSE) between the output and the training label while considering a penalty term to account for the constraint in equation (\ref{eq:opt_fun_constraint1}). This penalty term ensures that the model's predictions satisfies the constraint, thus aligning this approach with the RRM problem.

It is important to note that the MSE between the output and the training label is related to the first term being minimized in equation (\ref{eq:opt_fun}). This means that our regression-based approach consistently maintains a connection to the core optimization objectives of the RRM task, not only during the generation of the training dataset but also throughout the CNN training process.

\section{Performance assessment}\label{sec:metrics}

In this section, we assess the model's performance and determine its ability to make accurate predictions when presented with unseen data. 


\subsection{Traditional ML metrics}\label{subsec:ml_metrics}

In this subsection,  we evaluate the performance of the CNN models using ML metrics. These metrics offer quantitative measures of model effectiveness across various tasks, such as classification and regression.  The choice of metric depends on the specific problem being addressed. For regression, we employ metrics like MSE, MAE, and R-squared (R2) to evaluate the model's predictive accuracy in estimating continuous numeric values. On the other hand, classification tasks rely on metrics such as accuracy, precision, recall, and F1-score to assess the model's ability to correctly categorize data into discrete classes.

The accuracy, for instance, is defined as
\begin{equation}\label{eq:acc}
    \eta = \frac{T_N + T_P}{T_N+F_P+T_P+F_N},
\end{equation}
for a binary classification problem with positive and negative classes, where $T_N$ and $T_P$ are the true negative and true positive, and $FN$ and $FP$ are the false negative and false positive.

When the dataset is imbalanced,  recall and balanced accuracy are most suitable to evaluate the model. The recall is a type of accuracy per class defined as
\begin{equation}\label{eq:recall}
    \sigma= \frac{T_P}{T_P+F_N},
\end{equation}

By averaging the recall for each class, we obtain the balanced accuracy which is defined as 

\begin{equation}\label{eq:bal_acc}
    \phi = \frac{1}{2} \left (\frac{T_P}{T_P+F_N} + \frac{T_N}{T_N+F_P} \right).
\end{equation}

\subsection{Proposed ML metric for RRM task}\label{subsec:ml_metric_prop}

Our main goal is to fulfill the capacity demand for each beam, prioritizing capacity compliance over the perfect prediction of payload configurations. In this regard, we introduce the concept of flexible accuracy per payload configuration, a metric inspired by the recall equation (\ref{eq:recall}). This flexible accuracy per payload configuration
\begin{equation}\label{eq:acc_new1}
    \theta_{ l} =  \frac{1}{B} \sum_{b=1}^B\frac{S}{M_l},
\end{equation}
evaluates the performance of flexible payload models, where $S$ is the number of instances in which the offered capacity in beam $b$ was sufficient, and $M_l$ is the number of samples is class $l$. As in equation (\ref{eq:bal_acc}), we obtain the average flexible accuracy or balanced flexible accuracy
\begin{equation}\label{eq:acc_new2}
    \bar{\theta} = \frac{1}{L} \sum_{l=1}^L \theta_{ l}.
\end{equation}

\subsection{System performance metrics}\label{subsec:system_metrics}

When evaluating the model in terms of the system performance, we are interested in finding a payload configuration that ensures that the offered capacity satisfies the requested capacity for each beam. The offered capacity $\mathbf{c}_m = [C_{1,m}, C_{2,m}, C_{B,m}]$ is obtained after acquiring the payload configuration using the ML model. We then use the normalized mean square error (NMSE) 
\begin{equation}
    \nu_{m} = \frac{\sum [(\mathbf{c}_m - \mathbf{r}_m)^2]}{\sum [(\mathbf{r}_m)^2]}, m = 1 \cdots M_{\rm test}
\end{equation}
to measure the similarity between the offered $\mathbf{c}_m$ and requested capacity $\mathbf{r}_m  = [R_{1,m}, R_{2,m}, R_{B,m}]$ for each sample $m = 1, \cdots M_{\rm test}$ in the test dataset. Then the average NMSE can be computed as
\begin{equation}
   \nu_{\rm avg} =  \frac{1}{M_{\rm test}} \sum_{m=1}^{M_{\rm test}} \nu_{m}.
\end{equation}

\section{Simulation Results}\label{sec:sim_res}


\subsection{Simulation Setup}
The center frequency is 19 GHz, the satellite is positioned at 13 E, and the satellite altitude is 35786 km. The merit figure is $G/T = 17$ dB/K. The number of beams in the system is $B = 10$ and the beam centres are at 

\begin{equation}
\small
\begin{split}
    \phi_{\rm lat} = [39.3, 42, 44.7, 47.4, 51, 53.7, 56.4, 39.5, 42.2, 49] \\
    \phi_{\rm long} =[-5.3, 0, 5.3, 10.6, -0.5, 6, 12.3, 11.4, 16.7, 17.4].
    \end{split}
 \end{equation}

 By varying bandwidth and power, as described in subsection \ref{subsec:link}, we obtained the capacity per beam options shown in Table \ref{tab:possible_config_10beams_9config}. 

\begin{table}[h]

\caption{Possible resource allocations in a beam}\label{tab:possible_config_10beams_9config}
\adjustbox{scale=0.86}{%
\centering
\begin{tabular}{l|l|l|l|l|l|l}
\hline

\rowcolor[HTML]{EFEFEF} 
 Index & $\text{BW}_b$  & $P_b$  & $\text{EIRP3dB}_b$    & $\text{CINR}_b$   & $\text{SE}_b$   & $C_b$     \\ 
\rowcolor[HTML]{EFEFEF}  & [MHz]  &  [dBW]  &  [dBW]   &    & [bps/Hz]    &  [Mbps]     \\ \hline
 1     & 150 & 10 & 49.93 & 6.6670  & 1.9246 & 288.6844 \\ \hline
2     & 250 & 10 & 49.93 & 4.4741  & 1.5187 & 379.6639 \\ \hline
3     & 500 & 10 & 49.93 & 1.4831  & 0.9650 & 482.5084 \\ \hline
4     & 150 & 12 & 51.94 & 8.6396  & 2.2897 & 343.4547 \\ \hline
5     & 250 & 12 & 51.94 & 6.4615  & 1.8865 & 471.6312 \\ \hline
6     & 500 & 12 & 51.94 & 3.4817  & 1.3350 & 667.4827 \\ \hline
7     & 150 & 14 & 55.94 & 12.4904 & 3.0025 & 450.3720 \\ \hline
8     & 250 & 14 & 55.94 & 10.3705 & 2.6101 & 652.5215 \\ \hline
9     & 500 & 14 & 55.94 & 7.4357  & 2.0668 & 1033.4   \\ \hline
\end{tabular}}
\end{table}

The dataset comprises $M = 30,000$ samples, with $70\%$ allocated for training and $15\%$ each for validation and testing. The stochastic gradient descent (SGD) optimizer with learning rate $\mu = 0.01$ is used to minimize the loss function.

\subsection{Model evaluation and system performance}



\begin{figure}[h]
          \centering
        	\includegraphics[ width=7.cm]{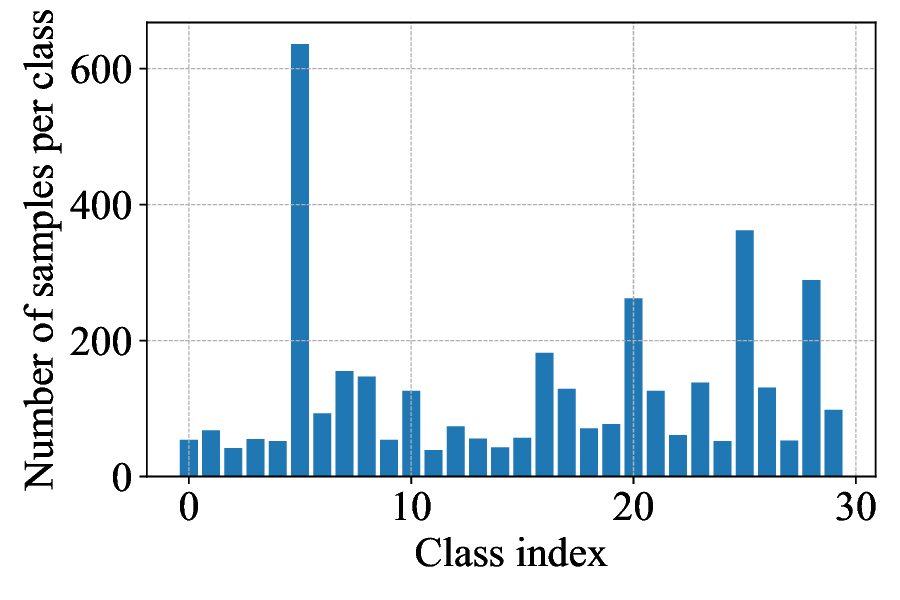}
        \caption{Number of samples per class in the validation set.}
        \label{fig:samp_per_class}
    \end{figure}

\begin{figure}[!h]
          \centering
        	\includegraphics[ width=7.cm]{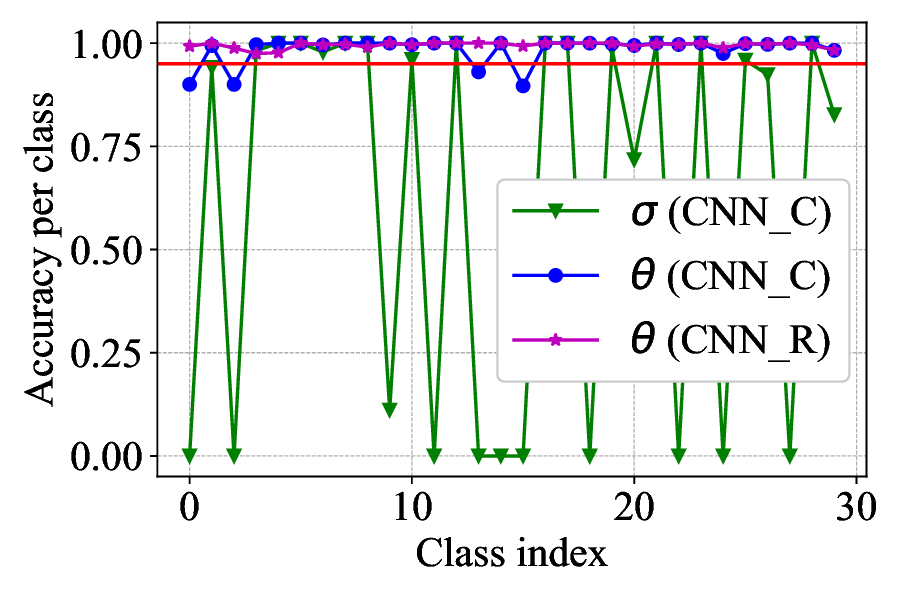}
        \caption{Proposed CNNs.}
        \label{fig:all_recall}
    \end{figure}

Due to the utilization of a realistic traffic model, certain payload configurations are more frequently generated within the system. Therefore, the dataset exhibits class imbalance, where some classes have a greater number of samples than others as shown in Figure~\ref{fig:samp_per_class}. 
In cases of imbalanced datasets, recall is the preferred metric for assessing the classification model's performance, as discussed in subsection \ref{subsec:ml_metrics}. Figure~\ref{fig:all_recall} presents the recall (green line) for each class when evaluating the ML-based flexible payload model using the classification approach (CNN\_C).

Classes with a higher number of data samples, such as classes 5 and 25, are corrected and classified by the model with an individual accuracy higher than $95 \%$, as indicated by the red line in Figure~\ref{fig:all_recall}. However, many classes are misclassified when considering CNN\_C.
Figure~\ref{fig:all_recall} also includes the flexible accuracy per class for CNN\_C (blue line). 
We can observe an improvement in terms of accuracy per class for CNN\_C with this new metric, but some classes still have an individual accuracy below $95 \%$.

\begin{figure}[!h]
          \centering
        	\includegraphics[ width=7.cm]{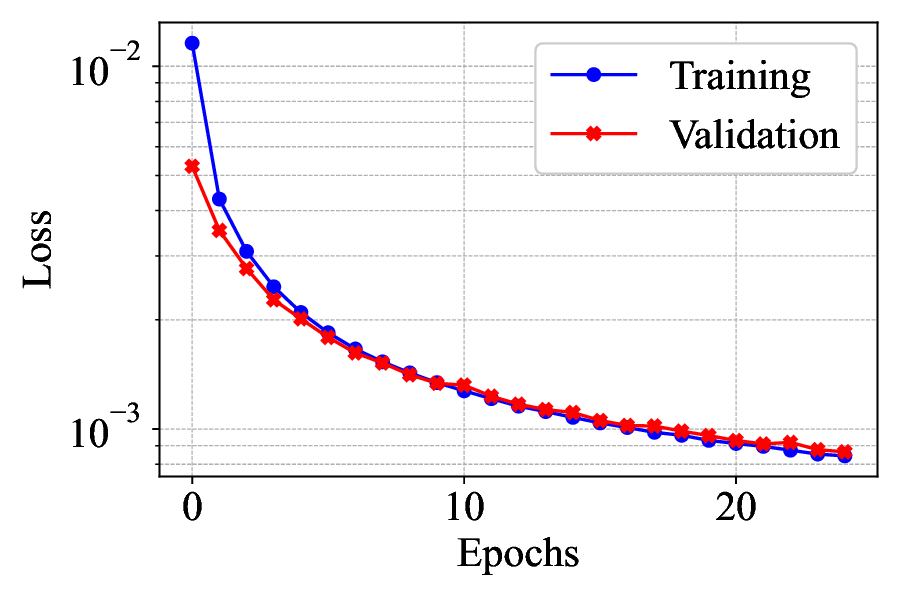}
        \caption{MSE obtained using the ML-based flexible payload model via regression.}
        \label{fig:cnn_regression_loss}
    \end{figure}

\begin{figure}[!h]
         \centering
        	\includegraphics[width=7.cm]{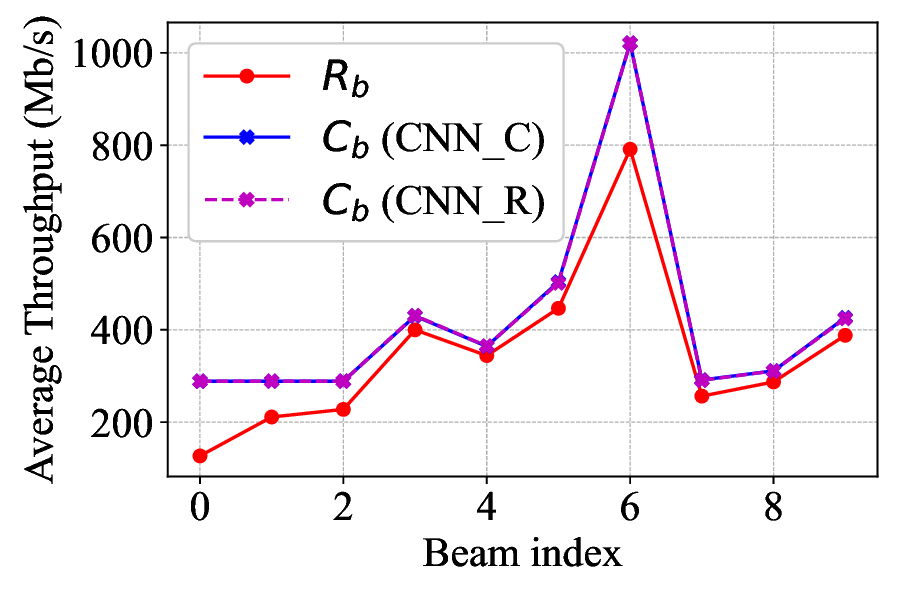}
        \caption{ Comparison between average offered capacity $C_b$ and requested capacity $R_b$.}
        \label{fig:cap_comp_10b}
    \end{figure}

The MSE results for the ML-based flexible payload model using regression are shown in Figure \ref{fig:cnn_regression_loss}. In contrast to CNN\_C, CNN\_R demonstrates satisfactory performance when evaluated with a traditional ML metric. Figure~\ref{fig:all_recall} presents the flexible accuracy per class (in magenta) for the ML-based flexible payload model via regression (CNN\_R). In such a case, all payload configurations achieve an accuracy higher than $95\%$.  This emphasizes the effectiveness of combining an appropriate evaluation metric with a model that intricately captures the RRM task.
The average offered capacity and requested capacity are compared in Figure~\ref{fig:cap_comp_10b}. On average, both models are able to satisfactorily obtain a payload configuration that leads to an offered capacity that satisfies the demand. In Table \ref{tab:ev_metrics}, we compare all the metrics used to evaluate the ML models. The traditional ML metrics used to evaluate the classification model CNN\_C failed to effectively capture the model's ability to meet the desired system benchmarks. On the other hand, when evaluating both the classification and regression models with the new ML metric, our expectations align more closely with the system's specific performance requirements.

\begin{table}[h]
\caption{Comparison between evaluation metrics}\label{tab:ev_metrics}
\adjustbox{scale=0.94}{%
\begin{tabular}{l|l|l}
\hline
\rowcolor[HTML]{EFEFEF} 
Type of Metric & Metric                                & Value    \\ \hline
ML             & Balanced accuracy for CNN\_C          & 61.28 \% \\ \hline
ML             & MSE + penalty term for CNN\_R                        & 0.00086  \\ \hline
ML + System    & Flexible balanced accuracy for CNN\_C & 98.50 \% \\ \hline
ML + System    & Flexible balanced accuracy for CNN\_R & 99.51 \% \\ \hline
System         & NMSE for CNN\_C                       & 0.0765   \\ \hline
System         & NMSE for CNN\_R                       & 0.0771   \\ \hline
\end{tabular}}
\end{table}

\section{Conclusions}\label{sec:concc}

In this work, we introduced a CNN to solve the RRM task considering flexible bandwidth and power. The RRM objective function and constraints were included in the ML loss function. We also proposed a new metric designed to balance traditional machine learning evaluation metrics and system performance. The simulation results indicate that traditional ML metrics fail to capture the system's requirements, whereas the proposed metric is robust to it.

\section*{Acknowledgment}

This work was supported by the European Space Agency (ESA) funded under Contract No. 4000134522/21/NL/FGL named “Satellite Signal Processing Techniques using a Commercial Off-The-Shelf AI Chipset (SPAICE)”. Please note that the views of the authors of this paper do not necessarily reflect the views of the ESA. Furthermore, this work was partially supported by the Luxembourg National Research Fund (FNR) under the project SmartSpace (C21/IS/16193290).

\ifCLASSOPTIONcaptionsoff
  \newpage
\fi



%
%
%

\bibliographystyle{IEEEtran}
\bibliography{refs}

%
%
%
%
%
%




\end{document}